\documentclass[11pt]{article}
\usepackage{tcolorbox}
\usepackage{subcaption}

\usepackage[preprint]{acl}
\usepackage{times}
\usepackage{latexsym}
\usepackage{booktabs}
\usepackage{csquotes}
\usepackage{amsmath} 
\usepackage{marvosym}
\usepackage{fontawesome}
\usepackage[T1]{fontenc}

\usepackage[utf8]{inputenc}

\usepackage{microtype}

\usepackage{inconsolata}

\usepackage{graphicx}

\title{{\ours}: An Auditable Multi-Agent System for Guideline-Grounded Diabetes Risk Screening}

\author{
 \textbf{Yung Wei Shueh\textsuperscript{1}\thanks{Equal contribution}},
 \textbf{Zhi-Jie Chen\textsuperscript{1}\footnotemark[1]},
 \textbf{Chia-Hsuan Hsu\textsuperscript{1}},
 \textbf{Hsin-Ling Hsu\textsuperscript{1}}, \\
 \textbf{Donghua Zhang\textsuperscript{4}},
 \textbf{Chenwei Wu\textsuperscript{4}}, 
 \textbf{Jun-En Ding\textsuperscript{2}},
 \textbf{Tongze Zhang\textsuperscript{3}}, 
 \textbf{Shihao 
 Yang\textsuperscript{2}}, 
 \textbf{Pengfei Hu\textsuperscript{3}},\\
 \textbf{Fang-Ming Hung\textsuperscript{1}},
 \textbf{Feng Liu\textsuperscript{2}}\thanks{Corresponding author: \texttt{feng.liu@rutgers.edu}}
\\
 \textsuperscript{1}Far Eastern Memorial Hospital, Taiwan 
 \textsuperscript{2}Rutgers University, USA \\
 \textsuperscript{3}Stevens Institute of Technology, USA
  \textsuperscript{4}University of Michigan, USA
}
\newcommand{\ours}{\textsc{DiaSentinel}}

\begin{document}
\maketitle
\begin{abstract}
Large language models (LLMs) offer promising clinical decision support but remain vulnerable to hallucinated facts, unsupported recommendations, and citation errors. We present DIASENTINEL, a fully on-premise multi-agent system for one-year type 2 diabetes mellitus (T2DM) risk screening and guideline-grounded report generation from electronic health records (EHRs). The system integrates calibrated risk prediction, deterministic clinical signal extraction, Reciprocal Rank Fusion over American Diabetes Association (ADA) guidelines, and a hybrid verification layer combining rule-based checks with LLM entailment. The demonstration provides a real-time batch-screening dashboard and an interactive patient report interface with cited recommendations, verification results, and raw EHR comparison. DIASENTINEL demonstrates a practical framework for reliable, auditable, and privacy-preserving LLM-based clinical decision support. Our system demonstration can be found at \url{https://diasentinel-demo.onrender.com/}
\end{abstract}

\section{Introduction}

Type~2 diabetes mellitus (T2DM) is among the most prevalent chronic diseases worldwide. According to the International Diabetes Federation, more than 500~million adults were living with diabetes in 2024, corresponding to approximately one in nine adults globally~\cite{idf2025}. T2DM is often preceded by an asymptomatic but modifiable high-risk period, during which early identification and intervention can delay or prevent disease onset and reduce subsequent complications~\cite{diabetes2002reduction,ada2026prevention}.

In clinical practice, routine clinical documentation is required for patients with potential chronic diseases, including measurements of glycated hemoglobin (HbA1c), fasting plasma glucose, body mass index (BMI), blood pressure, and lipid profiles. Concurrently, an effective risk indicator is needed to assess the near-term risk of type 2 diabetes mellitus (T2DM). Despite the significant advances of LLMs in the medical domain and EHR-based diagnosis in recent years~\cite{griot2025implementation,ding2024large}, most models have focused on algorithmic improvements without practical clinical deployment and application. In particular, LLMs introduce three major risks in medical applications. First, their predicted probabilities may be poorly calibrated and thus fail to reflect the actually observed disease incidence. Second, the model may hallucinate~\cite{asgari2025framework,kim2025medical} laboratory values, clinical findings, or recommendations that have no basis in the patient's medical record or clinical guidelines. Third, retrieval-augmented generation (RAG)~\cite{wu2025automated} may still associate otherwise valid recommendations with incorrect sources, thereby leading to \emph{citation drift}. These limitations reduce the interpretability, verifiability, and clinical reliability of LLM-based decision support.

To better reflect real-world clinical practice, we design a deployable, end-to-end LLM-based clinical decision support system, {\ours}, that decomposes the diagnostic workflow into a sequence of auditable, evidence-grounded subtasks. This modular design not only improves transparency and traceability but also enables more efficient retrieval and reasoning within each system component. More specifically, our contributions are threefold: (1) we develop a calibrated LLM-based risk predictor by fine-tuning Qwen2.5-14B~\cite{yang2024qwen25} with low-rank adaptation (LoRA)~\cite{hu2022lora}, aligning its estimated probabilities with the observed one-year T2DM incidence and supporting clinically meaningful risk stratification; (2) we introduce a two-stage retrieval pipeline over the American Diabetes Association (ADA) \textit{Standards of Care in Diabetes} guideline corpus that combines dense retrieval and cross-encoder reranking through Reciprocal Rank Fusion (RRF)~\cite{cormack2009reciprocal}, mitigating the length bias observed when the reranker is used alone; and (3) we integrate an in-the-loop verification agent that combines four deterministic checks with an LLM-based entailment check to audit factual consistency, unsupported content, and citation drift before the report is presented to clinicians.

Following this design, we present \textbf{\ours}\footnote{Live demo: \url{https://diasentinel-demo.onrender.com/}. The production system is deployed on hospital premises, while the public demo uses only synthetic patient data to protect privacy and demonstrate system functionality.}\footnote{Demo video: \url{https://youtu.be/yElcVSZXJHY}}, a fully on-premise multi-agent clinical decision support system for batch screening of one-year new-onset T2DM risk and on-demand generation of individualized, guideline-grounded reports. Implemented with LangGraph, {\ours} applies LLMs to risk estimation, report generation, and entailment-based verification, while all other components use deterministic rules or retrieval-only methods.

\begin{figure*}[t]    \centering    \includegraphics[width=0.8\textwidth]{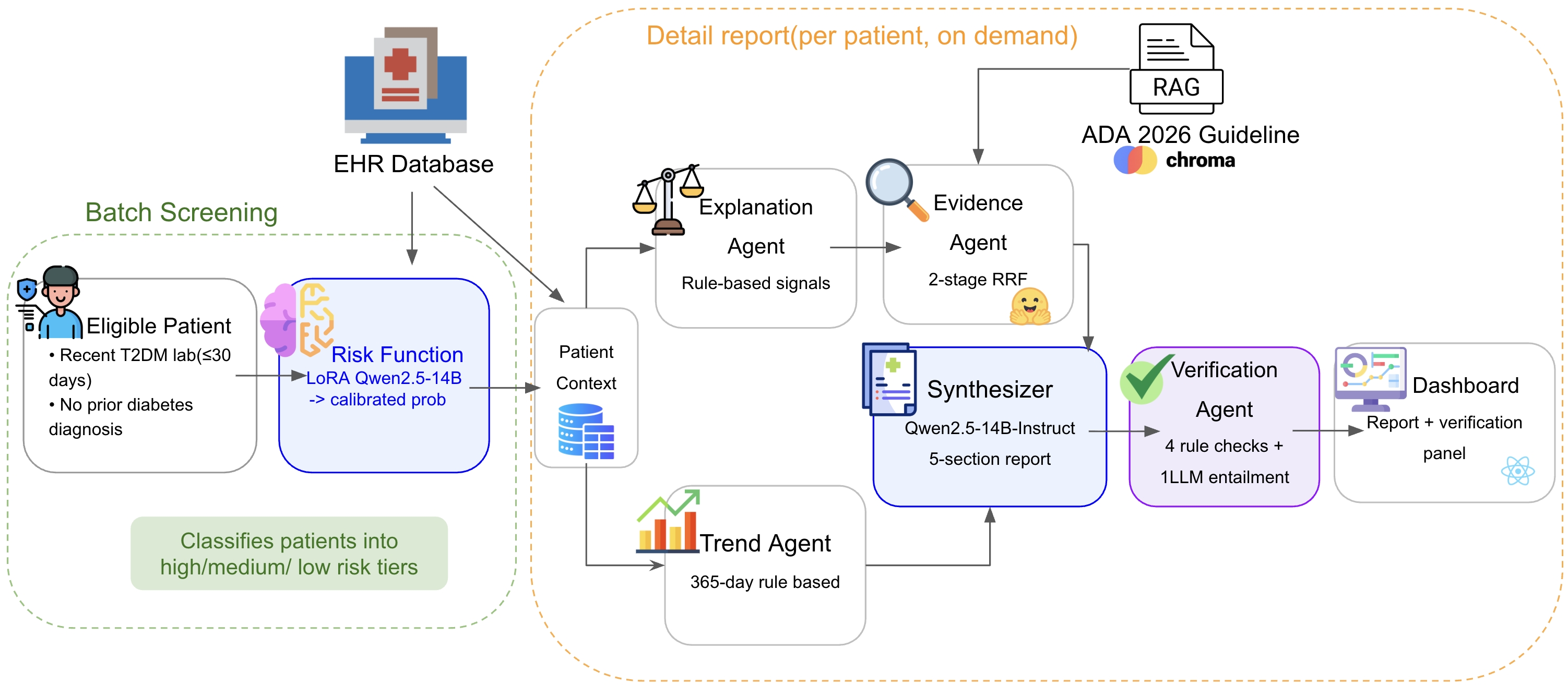}    \caption{The pipeline architecture. Blue: LLM-based; gray: deterministic; purple: hybrid.}    \label{fig:main_flow}\end{figure*} 

\section{{\ours} System Architecture}
\label{sec:system-components}

As shown in Figure~\ref{fig:main_flow}, DIASENTINEL is an on-premise clinical decision support pipeline orchestrated with LangGraph~\cite{duan2024exploration}. It separates population-level screening from patient-specific report generation through two modes:
\vspace{-2mm}
\begin{itemize}
\item \textbf{Batch risk screening.} The \textbf{Risk Function} assigns each eligible patient a calibrated one-year T2DM risk probability and a high-, medium-, or low-risk tier.
\vspace{-2mm}
\item \textbf{Detailed report generation.} Given a clinician-selected patient, the system retrieves longitudinal EHR records, extracts clinical signals, retrieves guideline evidence, synthesizes a structured report, and verifies it before display.
\end{itemize}

In this system architecture, {\ours} applies LLMs only to prediction, report synthesis, and semantic verification, while keeping EHR retrieval, factual extraction, numerical comparison, and evidence retrieval deterministic. Specifically, the LLM-based modules include the \textbf{Risk Function}, the \textbf{Synthesizer}, and the entailment check in the \textbf{Verification} layer. For screening, the Risk Function uses a LoRA fine-tuned Qwen2.5-14B model~\cite{yang2024qwen25} to estimate one-year new-onset T2DM risk from structured EHR-derived features. Token-level log probabilities are converted into a raw risk estimate via softmax and calibrated with Platt scaling~\cite{platt1999probabilistic}. The calibrated score is cached and reused across the dashboard, report summary, and audit checks to ensure consistency.

\subsection{Deterministic Clinical Tracking}
\label{sec:ehr-context-signals}

\paragraph{Real-Time Retrieval System.} To support real-time diagnosis, the physician first retrieves the patient's historical medical records from the hospital database via SQLAlchemy\footnote{\url{https://docs.sqlalchemy.org/en/20/}}. The retrieved context comprises SOAP (Subjective, Objective, Assessment, Plan) records~\cite{sorgente2005soap}, lab test results, and nursing assessments collected within a one-year time window. Next, we design an \textbf{Explanation Agent} to extract clinically interpretable risk signals using six deterministic threshold rules. These rules cover HbA1c, fasting glucose, BMI, blood pressure, low-density lipoprotein (LDL) cholesterol, and a metabolic-syndrome-related lipid pattern defined by elevated triglycerides with low high-density lipoprotein (HDL). When a rule is triggered, the agent emits a structured signal containing the variable, observed value, severity level, and a concise rationale, such as \enquote{An HbA$_{1c}$ value of 6.1\% falls within the prediabetes range}. The resulting signals are then ranked by clinical priority so that the Synthesizer can present the most relevant risk factors first.

\vspace{-2.5mm}
\paragraph{Longitudinal Trend Tracking Annotation.} In parallel, the \textbf{Trend Agent} summarizes longitudinal changes over the same 365-day window. For each monitored variable, it compares the earliest and most recent values using variable-specific noise-tolerance bands, assigns a trend label such as \texttt{stable}, \texttt{rising}, \texttt{improving}, or \texttt{insufficient\_data}, and records whether a clinically meaningful cut point has been crossed. The agent outputs both structured trend facts and a short natural-language summary, denoted as \texttt{summary\_nl}. The Synthesizer is instructed to reuse this summary verbatim rather than re-inferring the trend from raw values, which further limits opportunities for numerical hallucination.

\subsection{Knowledge-based Guideline Retrieval}
\label{sec:retrieval-synthesis}

To ground report generation in clinical guidelines, the \textbf{Evidence Agent} builds a local retrieval corpus from the ADA Standards of Care in Diabetes (2026). The guideline is split into 382 section-aware chunks (512 tokens max) using Docling~\citep{auer2024docling}, embedded with bge-m3~\citep{chen2024bge}, and indexed in Chroma with source, section, and page metadata. At inference, the agent retrieves dense candidates, reranks them with bge-reranker-v2-m3~\citep{chen2024bge}, and fuses both rankings via RRF to obtain the final \texttt{evidence\_chunks}; if reranking is unavailable, dense retrieval is used instead.

The \textbf{Synthesizer} (Qwen2.5-14B-Instruct) generates a five-part clinical report. Each recommendation is explicitly bound to retrieved metadata (source, section, and page), while unsupported statements are rendered as generic, non-cited guidance. This metadata-driven design prevents fabricated citations and grounds all referenced recommendations in the retrieved ADA evidence.

\vspace{-2mm}
\subsection{Clinically Reliable Verification}
\label{sec:verification-layer}

Before the report is presented to clinicians, the \textbf{Verification Agent} validates it against upstream structured outputs. Four deterministic checks ensure consistency between the report and (i) predicted risk score and tier, (ii) EHR-derived numerical values, (iii) longitudinal findings, and (iv) retrieved guideline citations, detecting common errors such as altered values, incorrect risk stratification, unsupported trends, and citation drift.

An additional Qwen2.5-14B-Instruct entailment check verifies whether recommendations are semantically supported by the retrieved guideline passages. Each check returns \texttt{pass}, \texttt{flag}, or \texttt{skipped}. The agent is strictly \emph{annotate-only}: it records all results in an append-only JSON Lines (JSONL) audit log and displays them to clinicians without modifying the generated report, preserving transparency and clinician authority.

\vspace{-2mm}
\section{Demonstration}
\label{sec:demonstration}

In this section, we demonstrate {\ours} through two interfaces: (1) a daily batch-screening dashboard and (2) a per-patient detailed report page. Our goal is to assist metabolic-care clinicians in more rapidly assessing patients' metabolic conditions and their risk of new-onset diabetes.

\vspace{-2mm}
\subsection{Daily batch-screening }

\begin{figure*}[!ht]
    \centering
    \begin{minipage}[c]{0.44\textwidth}   
        \centering
        \includegraphics[width=\linewidth]{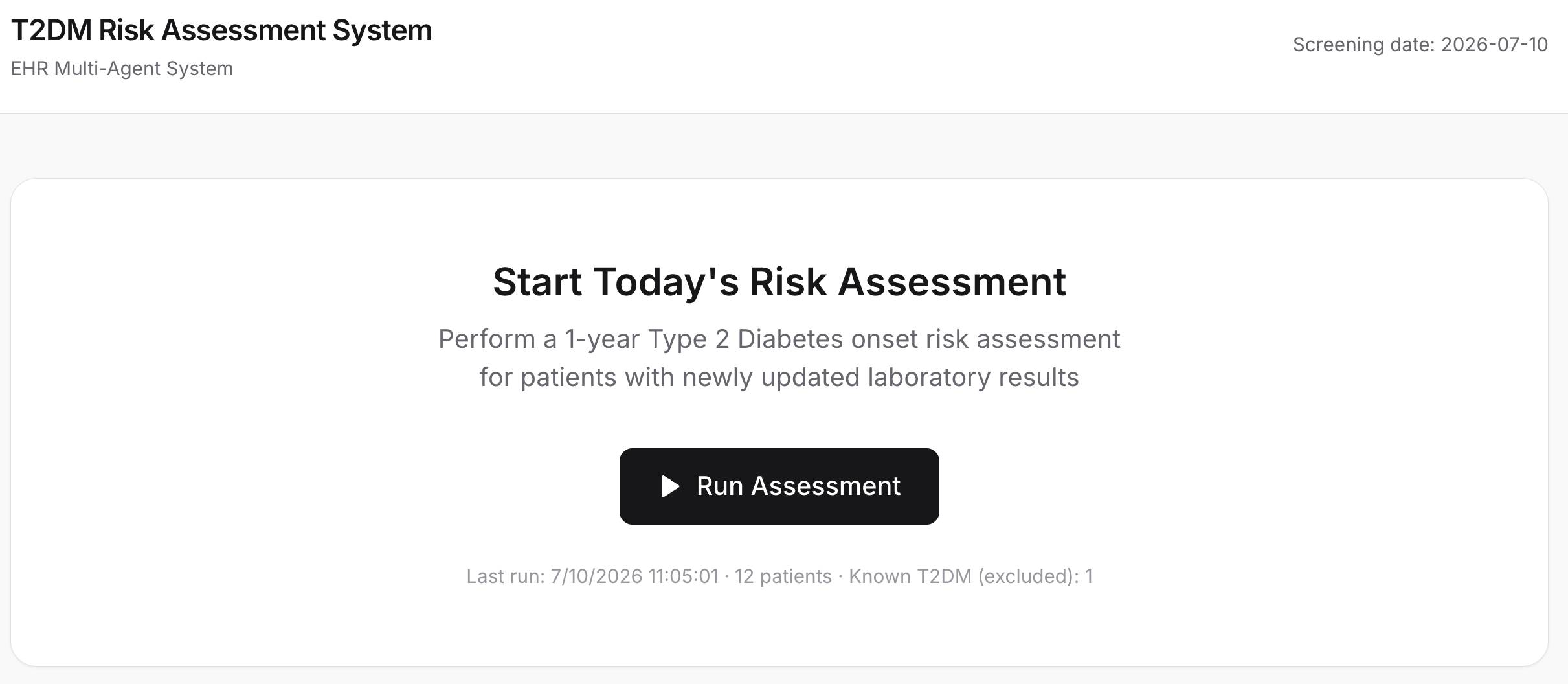}\\[2pt]
        \small (a)\par
        \vspace{1.2em}
        \includegraphics[width=\linewidth]{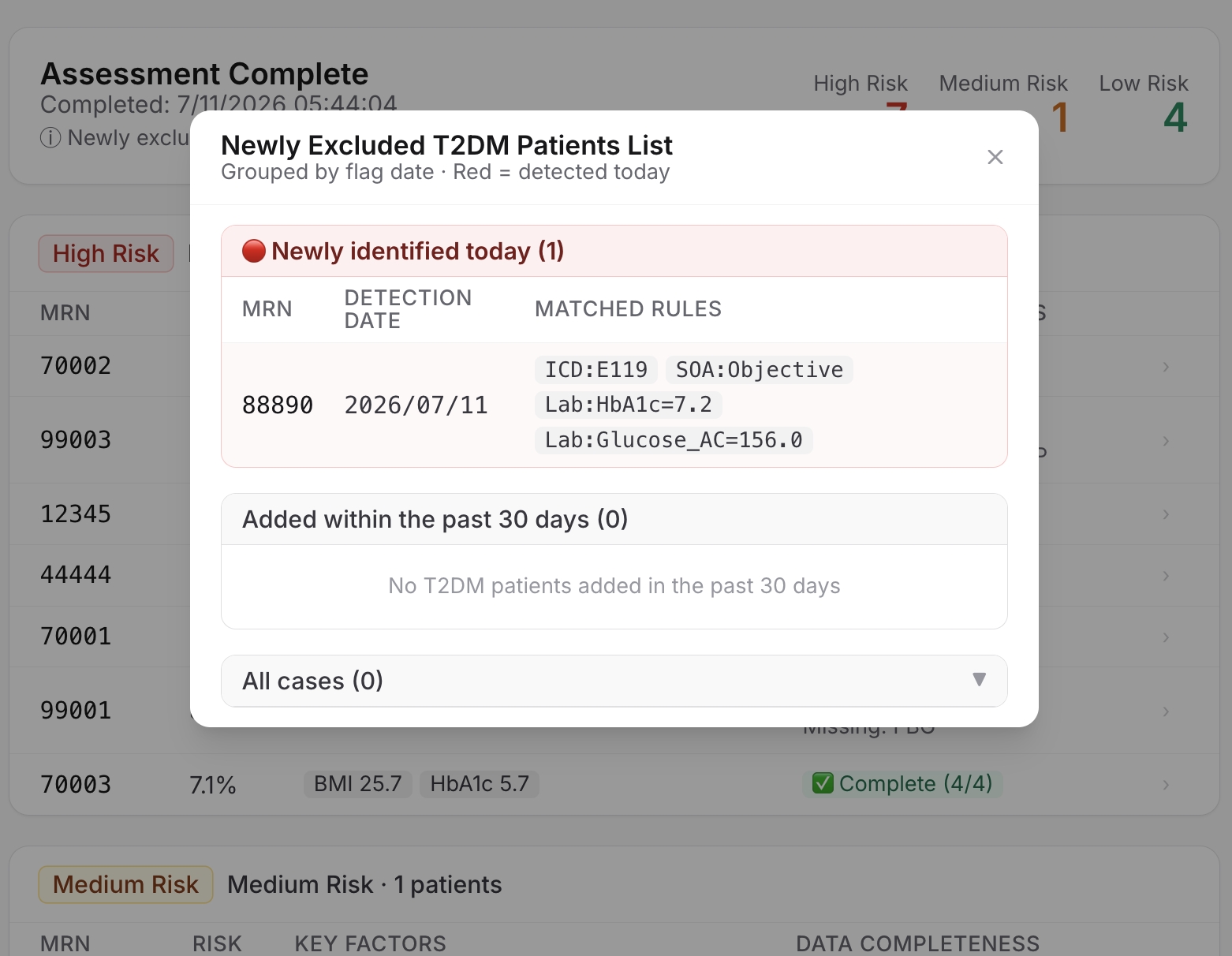}\\[2pt]  
        \small (b)
    \end{minipage}
    \hfill
    \begin{minipage}[c]{0.52\textwidth}   
        \centering
        \includegraphics[width=\linewidth]{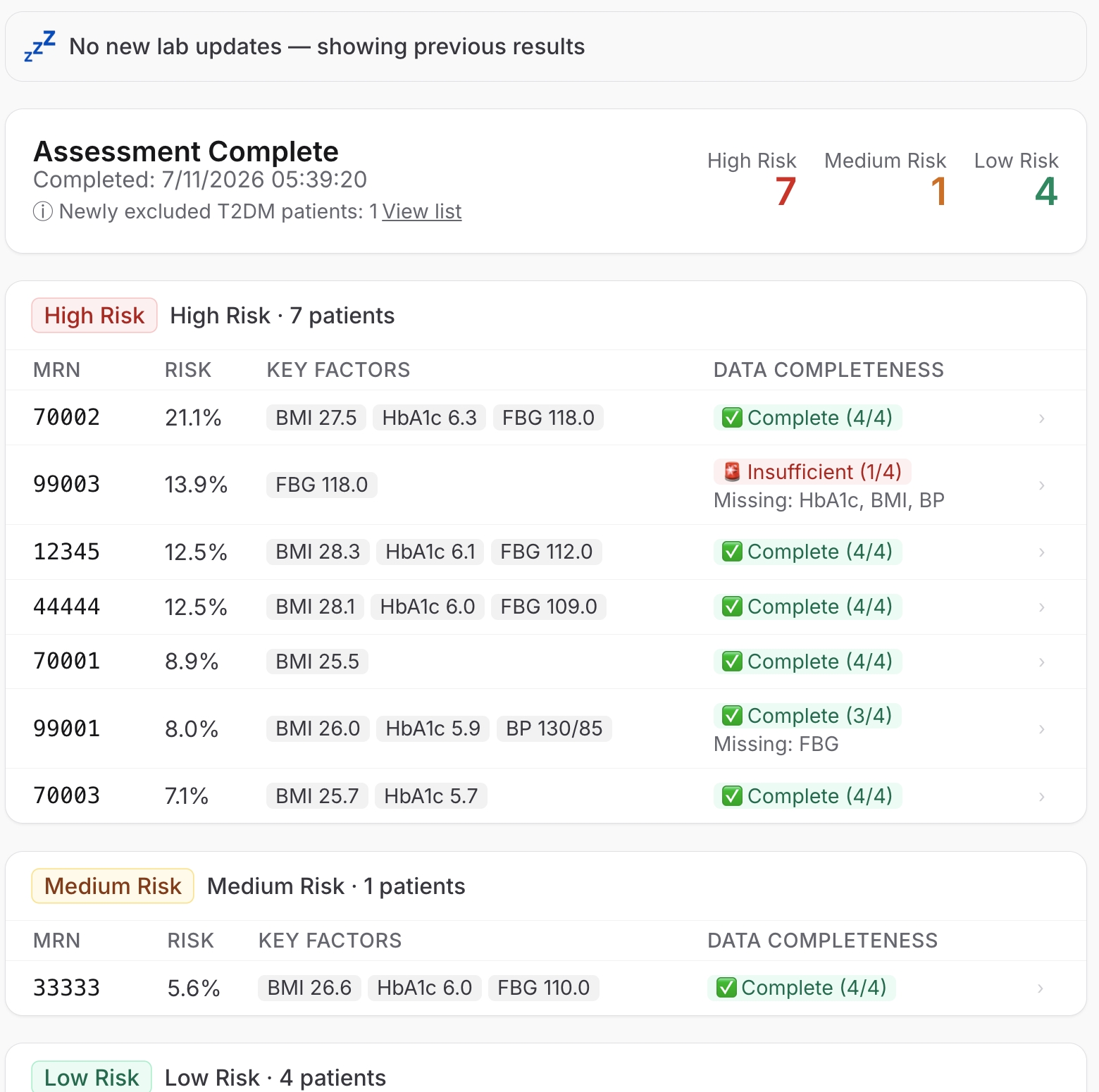}\\[2pt]
        \small (c)
    \end{minipage}
    \hfill

    \caption{The batch-screening interface shows \textbf{(a)} one-click assessment and run statistics, \textbf{(b)} an audit panel for suspected new-onset T2DM cases, and \textbf{(c)} a calibrated risk dashboard with data-completeness and screening-status indicators.}
    \label{fig:dashboard}
\end{figure*}

As illustrated in Figure~\ref{fig:dashboard}, we developed a clinician-facing tracking dashboard for batch patient screening (Figure~\ref{fig:dashboard} (a)) and risk monitoring (Figure~\ref{fig:dashboard} (c)). Clinicians can initiate screening by clicking the ``Run Screening'' button, after which the system evaluates all eligible patients in batch. The frontend displays real-time progress via server-sent events (SSE), while the backend sequentially performs risk prediction and data-completeness assessment for each patient. After screening, the dashboard stratifies patients into high-, medium-, and low-risk groups.

More specifically, each patient receives a data-completeness indicator reflecting input-data quality. This score is derived from four key clinical measurements, namely HbA1c, fasting glucose, BMI, and blood pressure, with each available measurement contributing one point, for a total of 0 to 4. Patients are labeled as high completeness (3--4 measurements), medium completeness (2), or incomplete (0--1). 

The dashboard also reports the execution status of each screening run. When new laboratory data are detected, it reports ``\textit{$N$ new patients were included in this screening run}''; otherwise, it displays ``\textit{No new laboratory data were detected; previous screening results were reused},'' reflecting the high-watermark-based idempotent skipping mechanism. Furthermore, patients with confirmed diabetes are excluded from the screening cohort. Because ICD-10 codes alone (E08--E13) are insufficient for confirmation, the system additionally reviews diagnostic notes, medication history, and abnormal glucose-related laboratory results. Cases meeting these criteria are flagged as potential newly developed T2DM and displayed on the frontend (Figure~\ref{fig:dashboard}(b)). This panel stratifies cases by flagging time into three views: ``\textit{newly identified today},'' ``\textit{added within the past 30 days},'' and ``\textit{all cases}.''

\subsection{Interactive Patient Reports }
\label{sec:detailed-report}

\begin{figure}[!ht]
    \centering
    \includegraphics[width=1\linewidth]{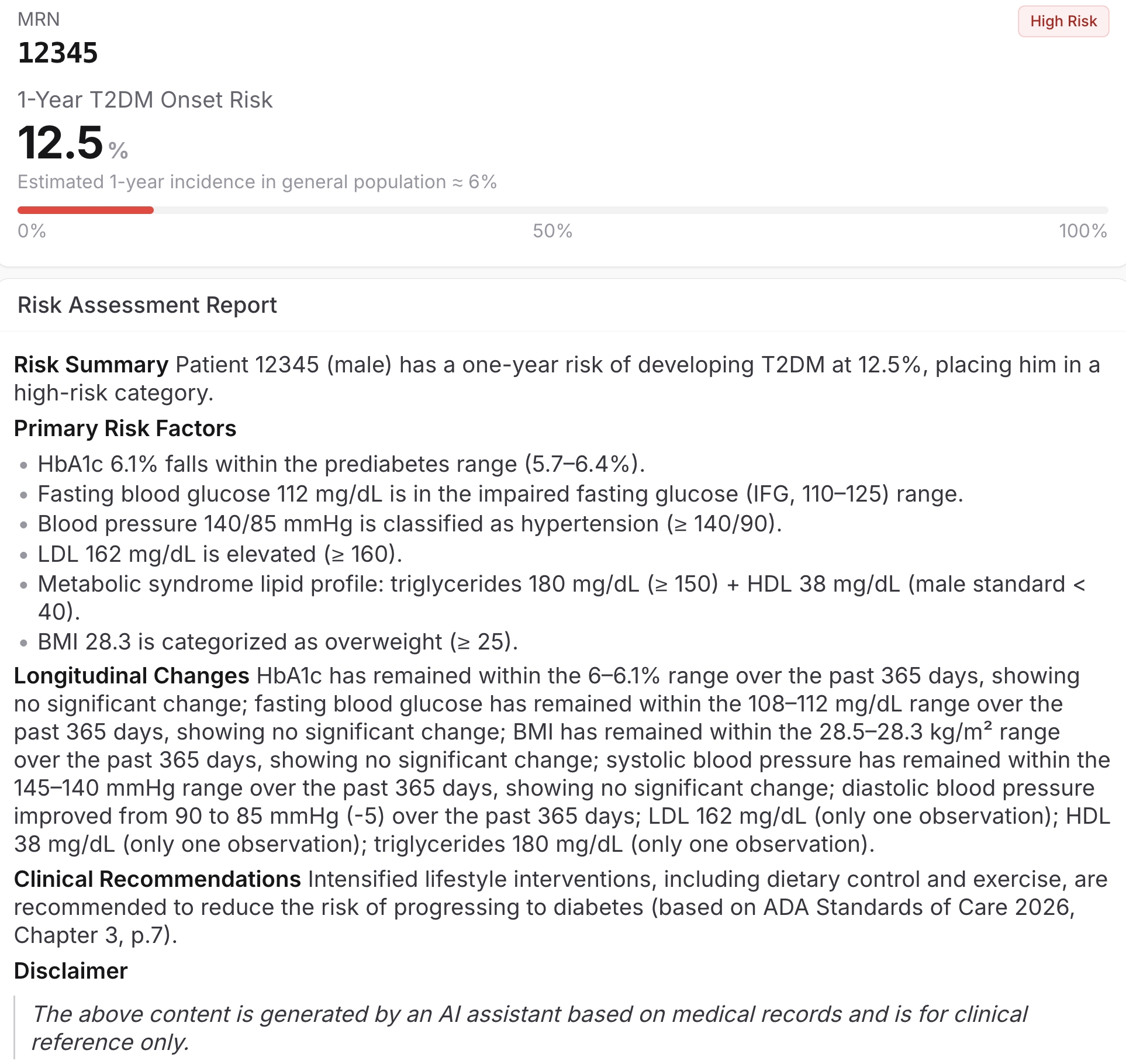}
    \caption{The five-part clinical report generated by {\ours}, including a risk summary, key risk factors, longitudinal trends, ADA-grounded recommendations with citations, and a disclaimer (see \S\ref{sec:detailed-report}).}
    \label{fig:report}
\end{figure}

\begin{figure}[!ht]
    \centering
    \includegraphics[width=\linewidth]{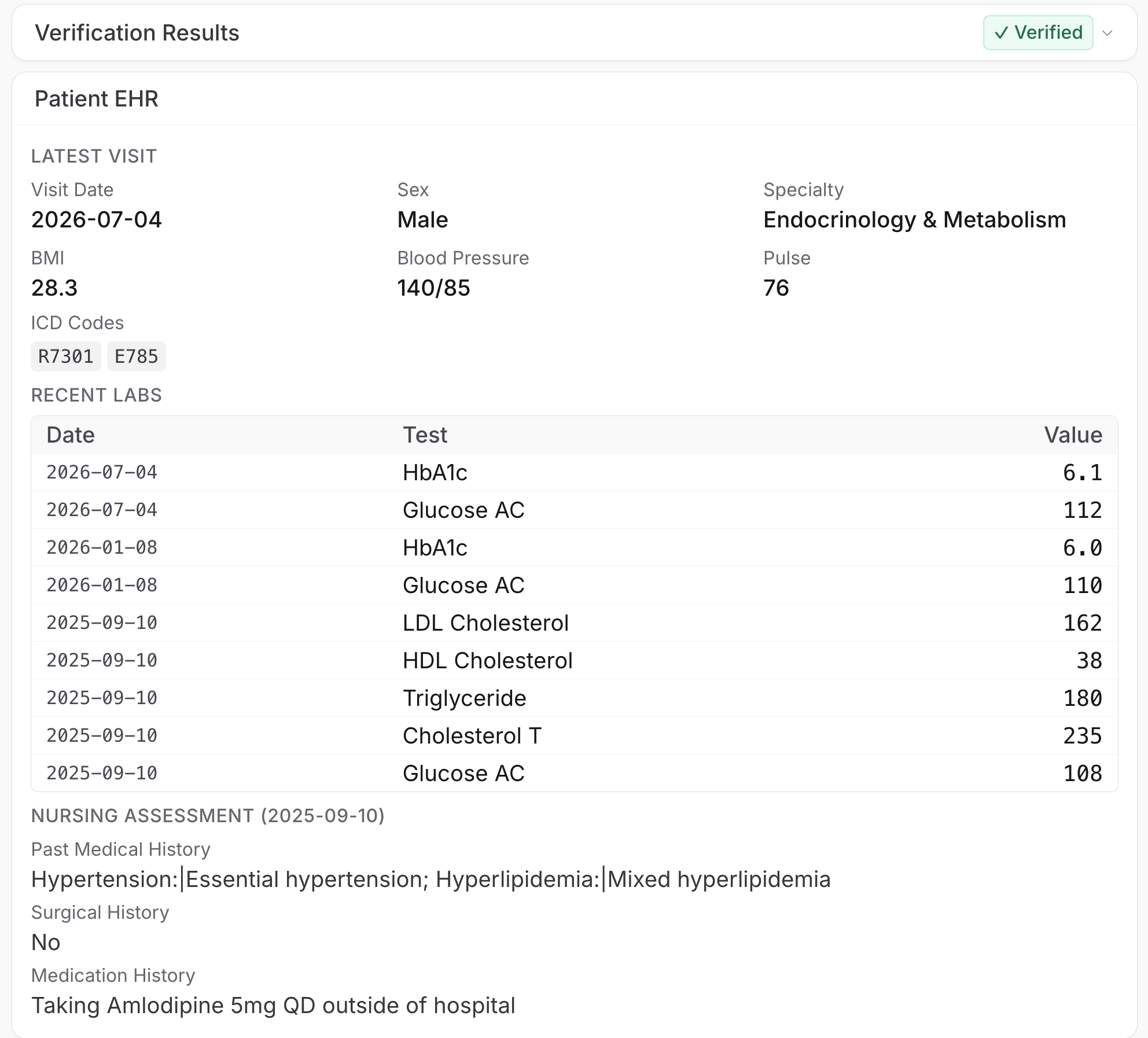}
    \caption{Lower half of the detail page: the verification-layer
    check-results panel, followed by the raw EHR tables}
    \label{fig:report_right}
\end{figure}

When a clinician selects a patient from any risk tier, the system opens a detail page presenting a complete clinical report. The top of the page shows the five-part report generated by {\ours} (Figure~\ref{fig:report}), followed by a panel reporting the verification-layer checks (Figure~\ref{fig:report_right}). Below it, the raw EHRs, including recent encounters, laboratory results, and nursing assessments, are shown in tabular form. Crucially, these records are fetched via direct SQL queries and never pass through an LLM, so the source data shown to clinicians is guaranteed to be faithful to the underlying database rather than paraphrased by the model.

As an example, Figure~\ref{fig:report} demonstrates \emph{Key Risk Factors}, which lists the signals extracted by the Explanation Agent, such as prediabetes, abnormal fasting glucose, and overweight, ordered by clinical urgency, while \emph{Longitudinal Changes} surfaces the Trend Agent's summary. 

\begin{tcolorbox}[colback=blue!5, colframe=blue!30, boxrule=0.5pt,
                  left=4pt, right=4pt, top=3pt, bottom=3pt]
\small\textbf{Sample case:} the patient \#12345 (male; BMI 28.3, HbA1c 6.1\%, fasting glucose 112, blood pressure 140/85, LDL 162).
\end{tcolorbox}

Furthermore, the remaining sections, such as \emph{Clinical Recommendations}, are grounded in the retrieved ADA guidelines. Each statement is annotated with its corresponding source and page number, and the report ends with a disclaimer. To keep the interface responsive, the detail page reuses the risk probabilities cached during batch screening and reruns only the downstream agents when necessary, generating the full report on demand.

\subsection{Verification Artifacts}
\label{sec:verification-artifacts}
On the detailed report page, the verification results appear in a
collapsible panel between the generated report and the raw EHR tables; the panel is collapsed when all checks
pass and expands automatically when any item is flagged. The panel
header summarizes the overall status (e.g., \textit{Verified} or
\textit{N issues flagged}); expanding it lists one row per check
finding across the five check types, each with its status and a
one-line rationale (the trend check reports one row per lab series).
For a flagged item, the interface additionally juxtaposes what the
report states with what the upstream facts support, letting the
physician inspect the discrepancy at a glance, as summarized below.

\begin{tcolorbox}[
    colback=gray!5,
    colframe=gray!40,
    boxrule=0.5pt,
    arc=2pt,
    left=6pt, right=6pt, top=6pt, bottom=6pt,
    title=\textbf{Human-in-the-Loop Verification},
    fonttitle=\small,
    coltitle=black,
    colbacktitle=gray!15
]
The system automatically flags potential errors, such as tampered values or incorrect citation sources, while the physician retains final authority over the diagnosis via the \textsc{Accept} and \textsc{Override} actions.
\label{tab:box2}
\end{tcolorbox}

\begin{figure}[!ht]
    \centering
    \includegraphics[width=\linewidth]{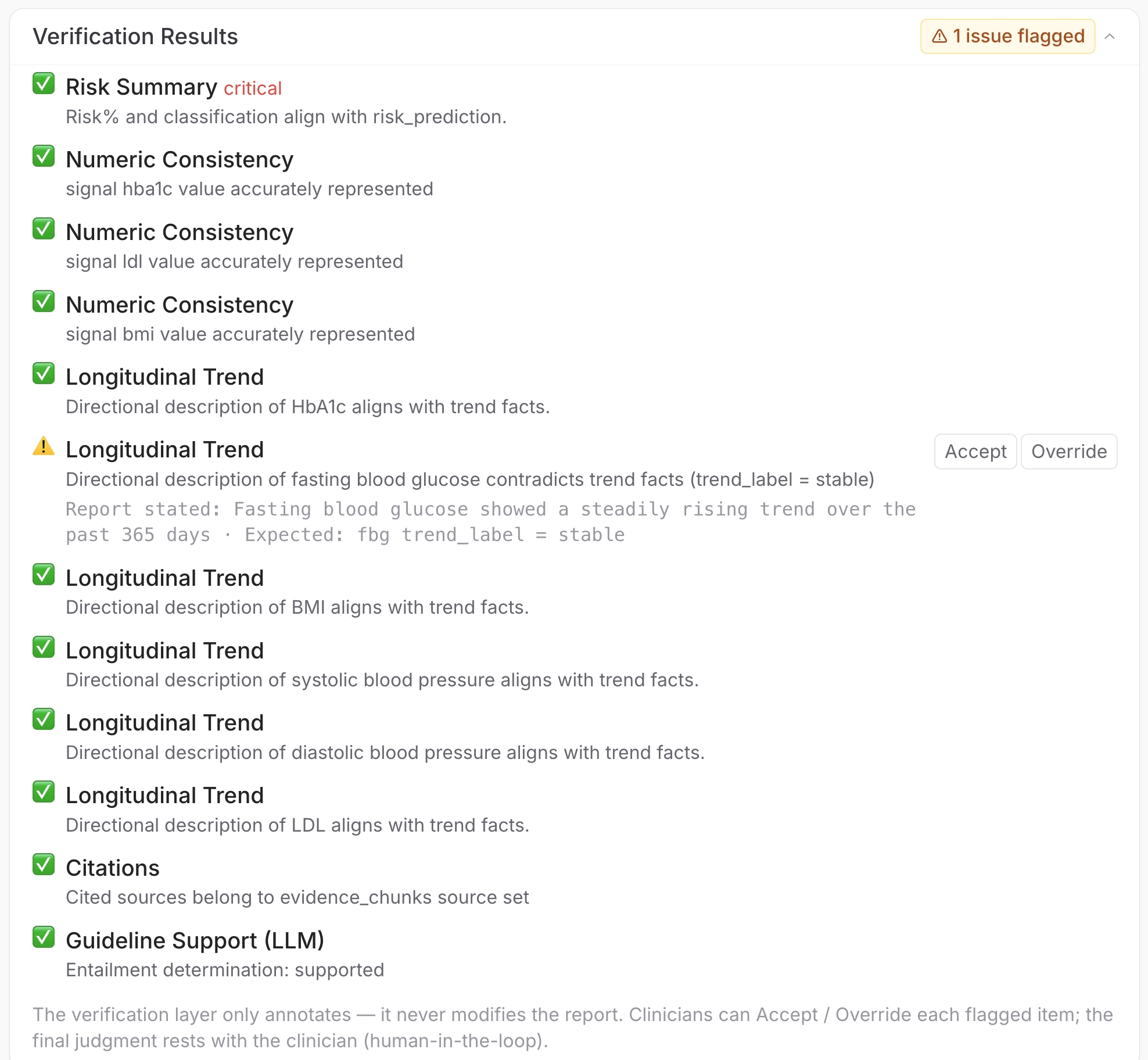}
    \caption{The flagged verification panel shows 11 of 12 checks passing, with the Longitudinal Trend discrepancy displayed alongside the expected value and \textsc{Accept}/\textsc{Override} controls.}
    \label{fig:verification}
\end{figure}

The panel in Figure~\ref{fig:verification} is itself produced by an
injected error, following the injected-error protocol summarized in Appendix
Table~\ref{tab:verification}: we tamper with an example report
(patient~\#70003) to claim that fasting blood glucose is rising,
whereas the structured trend fact records it as stable.

\section{System Evaluation}
\label{sec:evaluation}

\subsection{Evaluation Metrics} We evaluate DIASENTINEL along three dimensions. For risk prediction and stratification, we report the area under the receiver operating characteristic curve (AUROC) and the area under the precision–recall curve (AUPRC), with particular emphasis on AUPRC for the low-prevalence new-onset T2DM outcome ($\approx$ 6\%), and assess calibration via the Brier score $\text{BS} = \frac{1}{N}\sum_{i=1}^{N}(p_i - o_i)^2$, where $p_i$ is the calibrated predicted risk and $o_i$ the observed outcome; at a fixed threshold we additionally report accuracy, precision/positive predictive value (PPV), recall/sensitivity (Se), specificity (Sp), negative predictive value (NPV), and F1-score, with the high-risk threshold selected via Youden's $J = \mathrm{Se} + \mathrm{Sp} - 1$ and the medium-risk threshold chosen under a minimum-sensitivity constraint to prioritize screening recall. For guideline retrieval, we report Recall@5 and mean reciprocal rank (MRR) at the chunk level, Chapter@5 at the chapter level, and Cohen's $\kappa = \frac{p_o - p_e}{1 - p_e}$ between two independent annotators (where $p_o$ is the observed and $p_e$ the chance-expected agreement) to assess evaluation-set reliability. For verification reliability, we report the false-alarm rate on clean reports, the interception rate on manually injected errors, and the stability of the LLM-based entailment check under repeated fixed-input runs.

\subsection{Performance of Risk Function}
\label{subsec:risk_perf}
\begin{figure}[!ht]
    \centering
    \includegraphics[width=0.5\linewidth]{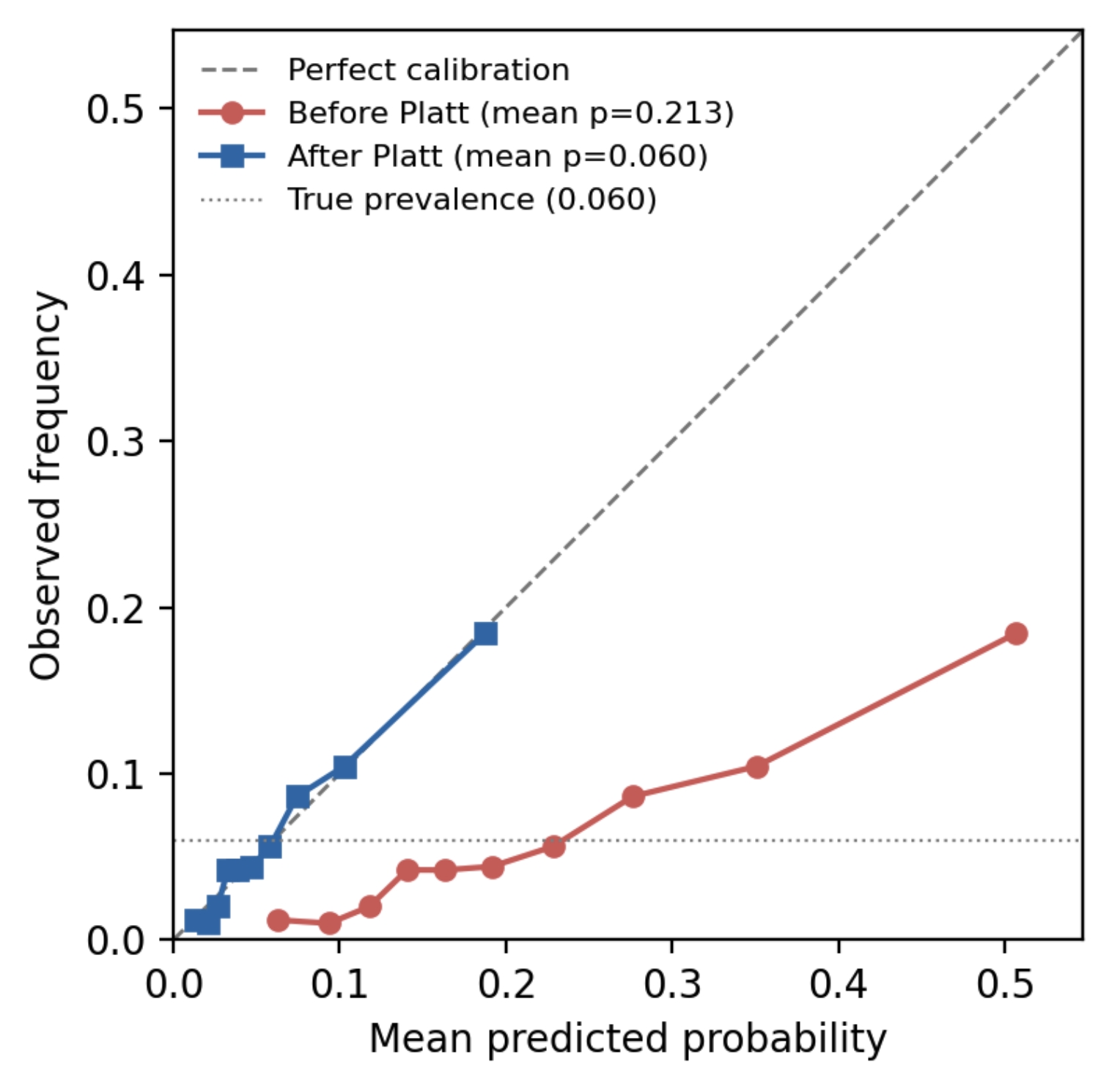}
    \caption{Reliability diagram of the risk predictor ($n=4{,}982$), showing that Platt scaling corrects overconfidence and reduces the mean predicted risk from 0.213 to 0.060, matching the observed one-year T2DM prevalence of 0.060.}
    \label{fig:calibration}
\end{figure}

To prevent the model output from exhibiting overconfidence ~\citep{guo2017calibration}, we apply Platt scaling ($a=1.0049$, $b=-1.5693$) to calibrate the predictions against the observed one-year disease incidence of approximately 6\% (Figure~\ref{fig:calibration}). The resulting calibrated probability serves as a unified risk score used for database storage, dashboard visualization, and automated report generation. From the validation set, we stratify patients into risk tiers based on thresholds for potential high-risk cases: high risk ($p \geq 0.056$), medium risk ($0.035 \leq p < 0.056$), and low risk ($p < 0.035$). Our results show that the high-risk threshold maximizes Youden's $J$, yielding 72\% sensitivity and 62\% specificity in the validation set, while the medium-risk threshold is selected as the value achieving at least 90\% sensitivity. This sensitivity-oriented design prioritizes the identification of potential high-risk cases while accepting additional false positives, consistent with the practical needs of clinical screening. Although these thresholds appear numerically low, in a low-prevalence setting they represent meaningful absolute risk, as the calibrated score aligns with the true incidence rate. Under a strict validation/test split, Risk Function achieves an AUROC of 0.737 (95\% confidence interval: 0.694--0.773) and a Brier score of 0.054, indicating effective discrimination and calibration. 
Next, we evaluate the held-out data: Table~\ref{tab:model_performance} reports the full held-out set ($N=4{,}982$) at the default threshold $p=0.5$, whereas Table~\ref{tab:risk_baselines} reports the test subset ($N=2{,}491$) with matched calibration. Both yield an AUROC of 0.737, with AUPRCs of 0.158 and 0.146, respectively, reflecting differences in cohort composition.

\begin{table}[!ht]
\centering
\caption{Comparison of Retrieval Strategies.}
\label{tab:retrieval_comparison}
\small
\setlength{\tabcolsep}{4pt}
\begin{tabular}{lccc}
\toprule
\textbf{Retrieval Strategy} & \textbf{Recall@5} & \textbf{MRR} & \textbf{Chapter@5} \\
\midrule
Vector (BGE-M3) & 0.704 & 0.659 & 0.918 \\
Vector $\rightarrow$ Rerank & 0.697 & 0.625 & 0.939 \\
\textbf{Vector + Rerank (RRF)} & \textbf{0.745} & \textbf{0.672} & \textbf{0.939} \\
\bottomrule
\end{tabular}
\end{table}

\begin{table}[!ht]
\caption{Overall Performance of the Proposed Model at the Default Threshold ($p=0.5$)}
\label{tab:model_performance}
\centering
\footnotesize
\begin{tabular}{lc}
\hline
\textbf{Metric} & \textbf{Value} \\
\hline
Accuracy & 0.8212 \\
Precision / Positive Predictive Value & 0.1623 \\
Recall / Sensitivity & 0.4733 \\
F1-score & 0.2417 \\
Specificity & 0.8434 \\
Negative Predictive Value & 0.9615 \\
AUROC & 0.7370 \\
AUPRC & 0.1582 \\
\hline
\end{tabular}
\end{table}

Beyond this global calibration, we verify that each deployed triage tier is itself well-calibrated. Table~\ref{tab:tier-calib}, computed on the same held-out split as the reliability diagram in Figure~\ref{fig:calibration}, reports the mean predicted probability and the observed one-year T2DM incidence within each risk tier. In all three tiers the predicted probability closely tracks the observed rate (High 0.115 vs.\ 0.117; Medium 0.048 vs.\ 0.052; Low 0.024 vs.\ 0.022), with the high-risk tier concentrating roughly twice the population prevalence ($\approx 0.060$). This tier-level calibration (not merely global calibration) is what makes the stratification directly actionable for triage: a patient placed in the high-risk bucket truly does carry an elevated, correctly-quantified one-year risk.

\paragraph{Comparison with Tabular Baselines}
To contextualize the Risk Function against standard tabular
predictors, we evaluate Logistic Regression and XGBoost under the
same patient-level split, feature window, calibration protocol, and
held-out test cohort. The proposed LLM achieves an AUROC of 0.737,
comparable to XGBoost (0.731) and higher than Logistic Regression
(0.697), with overlapping 95\% confidence intervals. XGBoost achieves
higher AUPRC (0.211 vs. 0.146), while the three models exhibit similar
Brier scores after identical Platt calibration (Table~\ref{tab:risk_baselines}).

\begin{table}[!ht]
\centering
\caption{Comparison with standard tabular baselines on the strict test set ($N=2{,}491$).}
\label{tab:risk_baselines}
\footnotesize
\setlength{\tabcolsep}{2.5pt}
\begin{tabular}{lcccccc}
\toprule
\textbf{Model} &
\textbf{AUROC} &
\textbf{AUPRC} &
\textbf{Brier} &
\textbf{Sens.} &
\textbf{Spec.} \\
\midrule
Logistic Regression & 0.697 & 0.130 & 0.057 & 0.673 & 0.648  \\
XGBoost             & 0.731 & \textbf{0.211} & \textbf{0.053} & 0.673 & \textbf{0.706}  \\
\textbf{$\star$ DIASENTINEL} & \textbf{0.737} & 0.146 & 0.054 & 0.660 & 0.704 \\
\bottomrule
\end{tabular}
\end{table}

\begin{table}[t]
\centering
\setlength{\tabcolsep}{6pt}
\small
\caption{Per-tier calibration on the held-out set ($n=4{,}982$). In every triage bucket the mean predicted probability closely matches the observed one-year T2DM incidence, confirming that the deployed operating thresholds deliver their promised risk levels (overall prevalence $\approx 0.060$).}
\label{tab:tier-calib}
\begin{tabular}{lrcc}
\toprule
\textbf{Risk Tier} & \textbf{n} & \textbf{Mean Pred.} & \textbf{Observed} \\
\midrule
High ($\geq 0.056$)   & 1619 & 0.115 & 0.117 \\
Medium ($\geq 0.035$) & 1184 & 0.048 & 0.052 \\
Low ($< 0.035$)       & 2179 & 0.024 & 0.022 \\
\bottomrule
\end{tabular}
\end{table}

\vspace{-2mm}
\subsection{Guideline Retrieval Performance}
\label{sec:eval-retrieval}

To evaluate the Evidence Agent's retrieval quality, we manually construct a 50-question evaluation set spanning five query types (high-level, boundary, abbreviation, negation, and trend), with 10 questions each. Of these, 49 are answerable from the corpus and 1 is a control question testing the system's ability to abstain. For each question, we annotate the golden chunk(s) from the corresponding ADA passage and record its chapter and page number (\texttt{golden\_citation}), yielding Cohen's $\kappa = 0.76$. RRF outperforms the dense-retrieval baseline on both Recall@5 and MRR, and achieves a chapter-level hit rate (Chapter@5) of 0.939.

The annotation schema was first validated on a 15-question pilot before being scaled to the full dataset. During the pilot, severity was reduced from three levels to two because the three-level scheme proved too subjective. We evaluate retrieval performance at both the chunk level using Recall@5 and MRR and the chapter level using Chapter@5. These metrics are used to compare different retrieval strategies (Table~\ref{tab:retrieval_comparison}). The main finding is described in Appendix~\ref{finding_result}.

\vspace{-2mm}
\subsection{Verification Agent Evaluation}
\label{sec:eval-verification}

As a demonstration-scale evaluation, we assess the Verification Agent on three aspects: false alarms on clean reports, detection of injected errors, and the stability and discrimination of the LLM-based entailment check. For each of the four deterministic checks, we construct three injected-error cases and three clean control cases, covering distinct error subtypes, magnitudes, and boundary conditions. Across the 24 cases, all 12 injected errors are successfully detected, while none of the 12 clean controls is falsely flagged. This corresponds to a sensitivity of 100\% and a specificity of 100\%.

For the LLM-based guideline-entailment check, we evaluate 20 grounded and 20 perturbed-unsupported claim--evidence pairs, each anchored to an ADA guideline chunk, achieving 80\% sensitivity and 100\% specificity (Appendix Table~\ref{tab:entailment-eval}). Across three runs at temperature~=~0, the judge yields identical verdicts for all 40 cases, indicating reproducible rather than stochastic behavior. Three of the four false negatives are conservative \emph{uncertain} judgments that our logic maps to pass, while the fourth results from numerical overlap between an unsupported claim and the retrieved evidence, highlighting a limitation of the judge.

These results offer a controlled assessment of the agent's ability to separate designed errors from clean cases and reject unsupported claims. However, the evaluation is limited to synthetic, manually constructed cases and does not reflect the error distribution or base rate of clinical deployment; it therefore does not establish real-world clinical reliability.

\section{Conclusion}
\label{sec:conclusion}
In this work, we demonstrated \textbf{DIASENTINEL}, an auditable on-premise LLM system for one-year T2DM risk screening and guideline-grounded clinical reporting. The system integrates calibrated risk prediction, evidence-grounded retrieval, and hybrid verification to improve the transparency and reliability of LLM-assisted clinical decision support while keeping all patient data within the hospital environment. Our system illustrates how these components work together through a real-time screening dashboard and interactive patient reports. 






\clearpage
\bibliography{custom}

\newpage

\appendix

\section{Appendix}
\label{sec:appendix}


\subsection{Model Development and Calibration}
\label{sec:eval-risk}
\begin{figure}[!t]
\centering 
\includegraphics[width=\linewidth]{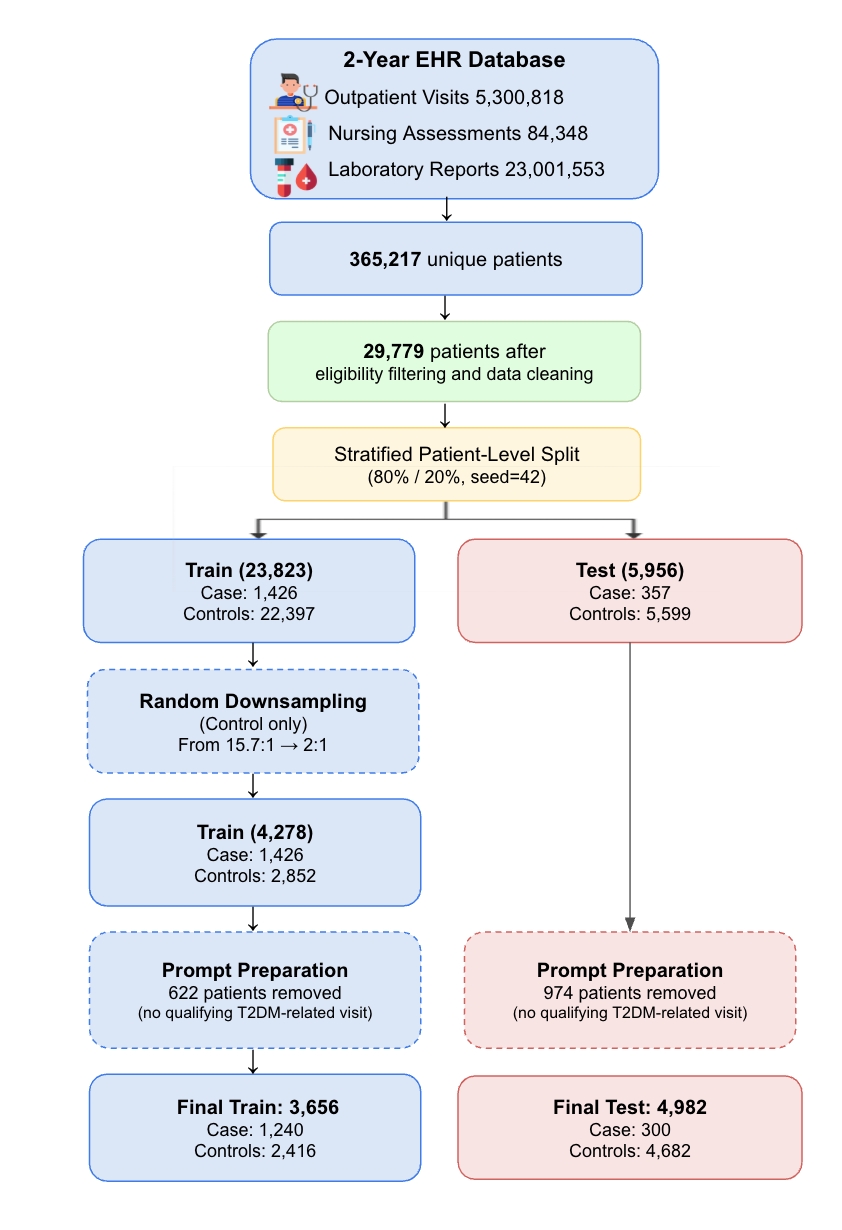} \caption{Dataset construction pipeline for model training and evaluation.} 
\label{fig:dataset-construction} 
\end{figure}

Figure~\ref{fig:dataset-construction} illustrates the dataset construction pipeline and the train/test split used for training and evaluating the Risk Function. The Risk Function is fine-tuned and calibrated on de-identified real-world patient EHR (private data) from a single hospital, attaining an AUROC of 0.737 (95\% CI [0.694, 0.773]) and a Brier score of 0.054 under a strict validation/test split, indicating good discrimination together with good probability calibration. Before calibration, the raw probabilities from the model's 2-token softmax are markedly overconfident: the mean predicted probability on the test set is about 0.213, whereas the true one-year T2DM incidence is only about 0.060. After Platt scaling, the predicted-probability distribution aligns much more closely with the observed prevalence (Figure~\ref{fig:calibration}), making the high/medium/low stratification better suited to clinical interpretation and use. This calibration step is especially important for \ours, because its downstream stages (dashboard risk stratification and report generation) all depend directly on this risk probability. Without proper calibration, the risk values could lead to over-alerting or mis-triage, degrading the quality of clinical decisions. Separately from this probability calibration, Table~\ref{tab:threshold_comparison} compares the classifier's default and optimized decision thresholds on the raw (pre-calibration) score, with the corresponding confusion matrix in Table~\ref{tab:confusion_matrix_optimized}. These are raw-score thresholds, not the calibrated screening thresholds (0.035/0.056) in \S\ref{subsec:risk_perf}.

\begin{table}[h]
\caption{Performance Comparison Between Default and Optimized Thresholds}
\label{tab:threshold_comparison}
\centering
\footnotesize
\setlength{\tabcolsep}{3pt}
\resizebox{\columnwidth}{!}{%
\begin{tabular}{lcc}
\hline
\textbf{Metric} & \textbf{Threshold = 0.5000} & \textbf{Threshold = 0.5622} \\
\hline
Accuracy & 0.8212 & 0.8418 \\
Precision / PPV & 0.1623 & 0.1747 \\
Recall / Sensitivity & 0.4733 & 0.4367 \\
F1-score & 0.2417 & 0.2495 \\
Specificity & 0.8434 & 0.8678 \\
\hline
\end{tabular}%
}
\end{table}

\begin{table}[h]
\caption{Confusion Matrix After Threshold Optimization}
\label{tab:confusion_matrix_optimized}
\centering
\footnotesize
\begin{tabular}{lcc}
\hline
\textbf{Actual Class} & \textbf{Predicted 0} & \textbf{Predicted 1} \\
\hline
Control (0) & 4063 & 619 \\
Case (1) & 169 & 131 \\
\hline
\end{tabular}
\end{table}

\subsection{Model Evaluation Results}
The clinical interpretation of the model performance metrics is summarized in Table~\ref{tab:clinical_interpretation} (full held-out set, $N=4{,}982$, at $p=0.5$, matching Table~\ref{tab:model_performance}).



\begin{table}[!t]
\caption{Clinical Interpretation of the Proposed LLM-Based Prediction Model}
\label{tab:clinical_interpretation}
\centering
\footnotesize
\begin{tabular}{p{2.3cm}p{5.0cm}}
\hline
\textbf{Aspect} & \textbf{Interpretation} \\
\hline
Discrimination ability & The model achieved an AUROC of 0.7370, indicating moderate discriminative ability. \\
Risk enrichment & The AUPRC of 0.1582 was higher than the positive class prevalence of 0.0602, suggesting enrichment of high-risk cases. \\
Sensitivity & The model detected 47.33\% of true incident diabetes cases at the default threshold. \\
Positive prediction & The PPV was 0.1623, indicating that most positive predictions were false positives. \\
Negative prediction & The NPV was 0.9615, suggesting relatively reliable negative predictions, partly due to class imbalance. \\
Clinical role & The model is more suitable for risk stratification and preliminary screening than for standalone diagnosis. \\
\hline
\end{tabular}
\end{table}

\begin{table}[t]
\centering
\caption{Performance of the deterministic verification checks on
clean and error-injected synthetic cases}
\label{tab:verification}
\footnotesize
\setlength{\tabcolsep}{5pt}
\begin{tabular}{lcccc}
\toprule
\textbf{Check} &
\textbf{Pos.} &
\textbf{Neg.} &
\textbf{Sens.} &
\textbf{Spec.} \\
\midrule
Risk summary        & 3 & 3 & 100\% & 100\% \\
Numerical agreement & 3 & 3 & 100\% & 100\% \\
Trend agreement     & 3 & 3 & 100\% & 100\% \\
Citation source     & 3 & 3 & 100\% & 100\% \\
\midrule
\textbf{Total}      & \textbf{12} & \textbf{12} &
\textbf{100\%} & \textbf{100\%} \\
\bottomrule
\end{tabular}
\end{table}

\begin{table}[t]
\centering
\caption{Performance of the guideline-entailment judge on synthetic
claim--evidence pairs}
\label{tab:entailment-eval}
\footnotesize
\setlength{\tabcolsep}{1.5pt}
\begin{tabular}{lccccccc}
\toprule
\textbf{Evaluation} &
\textbf{Pos.} &
\textbf{Neg.} &
\textbf{TP} &
\textbf{FP} &
\textbf{FN} &
\textbf{TN} &
\textbf{Sens. / Spec.} \\
\midrule
Guideline entailment &
20 & 20 & 16 & 0 & 4 & 20 &
80\% / 100\% \\
\bottomrule
\end{tabular}
\end{table}


\subsection{Key finding}\label{finding_result} On this clinical-guideline corpus, a cross-encoder reranker used alone \emph{degrades} ranking quality: its MRR is only 0.625, a drop of 0.034 below the baseline. We observe that the reranker tends to favor longer, narrative passages, whereas the content that actually contains the answer is often a shorter recommendation sentence or a table-like chunk, producing a verbosity bias. Per-question analysis shows only 2 questions are missed, neither solvable by the retrieval algorithm itself: one from a semantic gap in the corpus linking obstructive sleep apnea (OSA) to diabetes, the other from post-chunking/OCR information loss leaving only an abbreviated fragment. The main bottleneck thus lies in corpus coverage and chunk quality, not the retriever, making RRF without added LLM-based reranking a reasonable design choice.

\section{Ethics and Limitations}
\label{sec:limitations}


\paragraph{Decision Support, Not Replacement}
{\ours} is positioned to assist, not replace, clinical decision-making. Accordingly, every report is required to carry a ``for clinical reference only'' disclaimer that clearly delimits the system's role. The verification layer is limited to flagging and recording potential issues; it neither regenerates content nor blocks report output, and the final clinical judgment and treatment decisions remain the responsibility of trained medical professionals. As observed in prior work on emergency-care decision support ~\cite{he2025intriage}, automated tools can improve efficiency and convenience but may also lead users toward over-reliance, weakening their independent judgment. To mitigate this risk, we deliberately keep human decision-making within the system's flow and avoid having the system directly provide drug prescriptions or specific dosage recommendations. Concretely, the interface lets physicians accept or override each verification result (\S\ref{sec:verification-artifacts}) as a practical human-in-the-loop mechanism, ensuring that final decision authority always rests with human professional judgment.



\paragraph{Fairness and subgroup robustness are not yet fully evaluated}
We report tier-level calibration (Table~\ref{tab:tier-calib}), confirming that each triage bucket is well-calibrated; however, \emph{demographic} subgroup calibration—e.g., whether calibration and error rates are consistent across sexes—remains future work. This is limited by the demographic variables available in the data (sex is the only demographic feature the model uses; no age or ethnicity), and we have likewise not yet evaluated robustness across clinical departments.

\section*{License}

The {\ours} production system and its source code are not released under a public software license. The publicly accessible demonstration uses only synthetic patient data and is provided solely for demonstrating the system's functionality.

\section*{Funding}

This work was supported by the Far Eastern Memorial Hospital Innovation Research Project (Grant No. FEMH-2023-007).

\end{document}